\documentclass[letterpaper, 10 pt, conference]{ieeeconf}  

\IEEEoverridecommandlockouts                              

\usepackage{amsmath} 
\usepackage{amssymb}  
\usepackage{algorithm,algorithmic}
\usepackage{nccmath}
\usepackage{color}
\usepackage{tikz}
\usepackage{tikz-3dplot}
\usepackage{pgfplots}
\usepackage{tikzscale}
\pgfplotsset{compat = 1.3}
\usepackage{hyperref}
\usepackage{units}
\usepackage[outline]{contour}

{\left\lbrace\begin{array}{@{}l@{}}}%
	{\end{array}\right.}

\newcommand{\bv}[1]{\boldsymbol{#1}}  
\newcommand{\bvt}[1]{\boldsymbol{\mathbf{#1}}}  
\newcommand{\e}[1]{\bv{e}_{#1}} 

\title{\LARGE \bf
Design, Optimal Guidance and Control of a Low-cost\\ Re-usable Electric Model Rocket
}

\author{Lukas Spannagl$^{\ddag}$, Elias Hampp$^{\dag}$, Andrea Carron$^{\ddag}$, Jerome Sieber$^{\ddag}$, Carlo Alberto Pascucci$^{\dag}$,\\
Aldo U. Zgraggen$^{\dag}$, Alexander Domahidi$^{\dag}$, Melanie N. Zeilinger$^{\ddag}$

\thanks{This work was partly supported by the European Space Agency (ESA) under the Future Launchers Preparatory Programme (FLPP) Contract No. 4000124652/18/F/JLV.}

\thanks{$^{\dag}$E. Hampp, C.A. Pascucci, A. U. Zgraggen, and A. Domahidi are with Embotech AG, 8005 Zurich, Switzerland
        {\tt\small \{hampp,pascucci,\!zgraggen,domahidi\!\}@embotech\!.\!com}
}
\thanks{$^{\ddag}$L. Spannagl, A. Carron, J. Sieber, and M. N. Zeilinger are with the Institute for Dynamic Systems and Control (IDSC), ETH Zurich, 8092 Zurich, Switzerland
        {\tt\small \{spalukas,carrona,jsieber,mzeilinger\}@ethz.ch}
}
}

\begin{document}

\maketitle
\thispagestyle{empty}
\pagestyle{empty}

\begin{abstract}
In the last decade, autonomous vertical take-off and landing (VTOL) vehicles have become increasingly important as they lower mission costs thanks to their re-usability. However, their development is complex, rendering even the basic experimental validation of the required advanced guidance and control (G\,\&\,C) algorithms prohibitively time-consuming and costly. In this paper, we present the design of an inexpensive small-scale VTOL platform that can be built from off-the-shelf components for less than 1000~USD. The vehicle design mimics the first stage of a reusable launcher, making it a perfect test-bed for G\,\&\,C algorithms. To control the vehicle during ascent and descent, we propose a real-time optimization-based G\,\&\,C algorithm. The key features are a real-time minimum fuel and free-final-time optimal guidance combined with an offset-free tracking model predictive position controller. The vehicle hardware design and the G\,\&\,C algorithm are experimentally validated both indoors and outdoor, showing reliable operation in a fully autonomous fashion with all computations done on-board and in real-time. 
\end{abstract}

\begin{keywords}
Aerial Systems: Mechanics and Control, Motion and Path Planning, Optimization and Optimal Control
\end{keywords}

\section{Introduction}
Sustainable and repeatable access to space requires drastic changes in the way space transportation systems are designed and operated. In this regard, novel approaches to guidance, navigation, and control are key enabling technologies as they allow launchers to be reused  by performing pin-point landings. This significantly lowers the cost of access to space and maximizes the science return for missions on other planetary bodies, where it is necessary to land close to relevant geological features. Concerning guidance and control (G\,\&\,C), this paper focuses on the use of numerical optimization to generate trajectories and actuation profiles on-board and in real-time. Similar techniques are already used by SpaceX~\cite{Blackmore2016} demonstrating their potential in the real-world. The last decade of advances in embedded computational hardware, numerical algorithms~\cite{Acikmese2007, Malyuta2019, Liu2014, Sanchez2018}, and optimization software~\cite{Domahidi2012, Jerez2014} has made optimization-based G\,\&\,C algorithms feasible for space applications. Despite the clear potential and the active research in this area~\cite{JGCDspecial,Eren2017}, these techniques also pose new challenges in terms of performance and functional validation.

\begin{figure}
  \begin{center}
    \includegraphics[width=\linewidth, trim={12cm 2.5cm 3cm 0},clip]{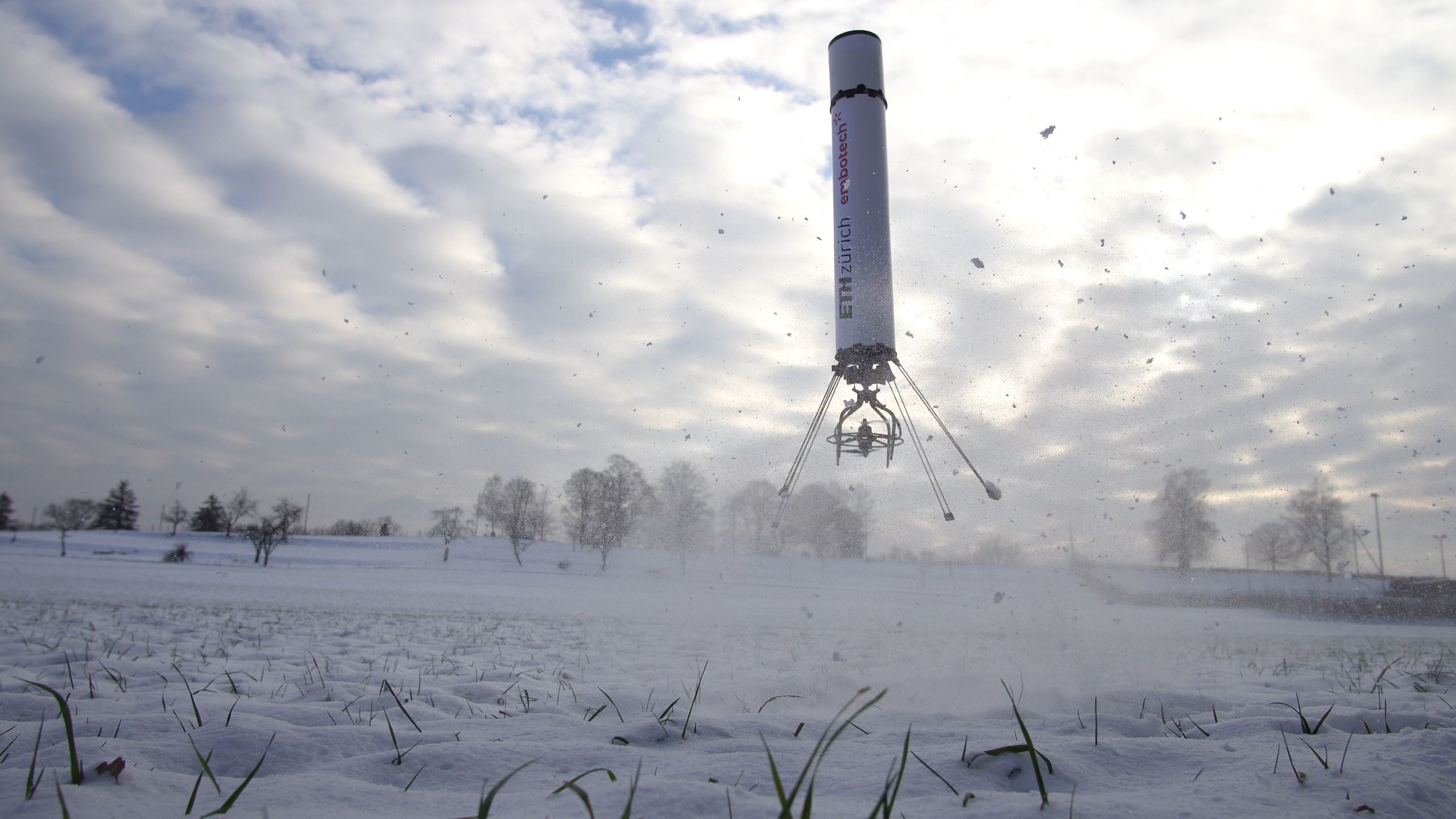}
  \end{center} 
  \caption{The EmboRockETH during an outdoor flight.}
   \label{fig:emborocket}
\end{figure}

In order to address these problems, we designed an inexpensive small-scale vertical take-off and landing (VTOL) aircraft, informally dubbed \emph{EmboRockETH} and shown in Figure~\ref{fig:emborocket}, which allows the analysis and validation of G\,\&\,C algorithms before their deployment on more complex systems. The EmboRockETH is designed to replicate some of the key aspects of a reusable rocket's first stage including aerodynamic control fins and a gimbaled main engine. Our contribution is twofold: we detail the complete hardware design of the EmboRockETH and we propose a G\,\&\,C pipeline for indoor and outdoor flights.\footnote{A video summarizing this paper and presenting footage of an outdoor experiment can be found here: \url{https://youtu.be/r4V7S1Rsbkg}}

The remainder of the paper is organized as follows. Next, we discuss existing related work and introduce the notation, before detailing the hardware design of the EmboRockETH in Section~\ref{sec:hw}. Subsequently, we outline the guidance and control architecture of the vehicle in Section~\ref{sec:control-architecture}, discuss experimental results in Section~\ref{sec:experiments}, and conclude the paper in Section~\ref{sec:conclusion}.

\subsection{Related Work}
In recent years, several test platforms similar to the EmboRockETH have been developed or are currently under development. The two most notable examples being the Grasshopper~\cite{grasshopper}, which was used by SpaceX for the early testing of their propulsive landing maneuvers, and Xombie from Masten Space Systems, which was used within the frame of the ADAPT project~\cite{scharf2017implementation} to test the lossless convexification-based GFOLD algorithm~\cite{acikmese2012g} and the terrain relative navigation system for the Mars 2020 mission~\cite{farley2020mars}. More recently, the European Space Agency has joined these efforts with the Future Launchers Preparatory Programme~(FLPP)~\cite{Preaud2019}. However, all the aforementioned vehicles are complex systems that require an appropriate budget, a crew and stringent safety measures for their operation. Hence, the use of quadrotors has been predominant in academic research, since these vehicles enable rapid prototyping and real-world testing of advanced G\,\&\,C methods~\cite{schoellig2012optimization, szmuk2018real} in an agile setting. Nevertheless, quadrotors share only a few basic principles with reusable first stages. Therefore, the vehicle presented in this paper is aimed at bridging this gap by providing an agile research platform for advanced G\,\&\,C methods with an improved dynamical representation of reusable launchers.

\subsection{Notation}
In this paper, we denote vectors with bold letters, i.e.,~$\mathbf{a} \in \mathbb{R}^n$, matrices with bold capital letters, i.e., $\mathbf{A} \in \mathbb{R}^{n \times m}$, and the time derivative of a vector or matrix with the dot notation, i.e.~$\dot{\mathbf{a}}$. The symbol $\mathbb{H}$ denotes the set of unit quaternions and $\e{i}$ denotes a unit vector along the $i$-th axis. We denote with the symbols~$\boldsymbol{p} \in \mathbb{R}^3$ and~$\boldsymbol{v} \in \mathbb{R}^3$ the position and velocity of the EmboRockETH in the world frame, respectively. The quaternion~$\boldsymbol{q} \in \mathbb{H}$ denotes the aircraft's attitude, while~$\boldsymbol{\omega} \in \mathbb{R}^3$ is the associated angular velocity. The symbol~$\mathbf{T} \in \mathbb{R}^3$ represents the thrust vector applied to the aircraft and~$\mathbf{g} \in \mathbb{R}^3$ denotes the gravitational force. Given a quaternion $\boldsymbol{q} \in \mathbb{H}$, the rotation matrix $\boldsymbol{R}(\boldsymbol{q})\in \mathbb{R}^{3\times 3}$ is defined as
\[
\boldsymbol{R}(\boldsymbol{q}) = \begin{medsize} \begin{bmatrix} 1-2 (q_y^2 + q_z^2)            & 2 (q_x  q_y-q_z  q_w) &  2 (q_x  q_z+q_y  q_w)\\
                                2 (q_x  q_y+ q_z  q_w) & 1-2 (q_x^2 + q_z^2)           &  2 (q_y  q_z-q_x  q_w)\\
                                2 (q_x  q_z-q_y  q_w)  & 2 (q_y  q_z+q_x  q_w) &  1-2 (q_x^2 + q_y^2) \end{bmatrix} 
\end{medsize},
\]
where $q_{x/y/z/w}$ are the components of the quaternion $\boldsymbol{q}$. Finally, given a quaternion $q \in \mathbb{H}$, we define the quaternion propagation matrix as 
\[
\boldsymbol{Q}(\boldsymbol{\omega}) = \begin{bmatrix} 0        & -\omega_x & -\omega_y & -\omega_z\\
                                                      \omega_x &  0        &  \omega_z & -\omega_y\\
                                                      \omega_y & -\omega_z &  0        &  \omega_x\\
                                                      \omega_z &  \omega_y & -\omega_x &  0 \end{bmatrix},
\]
where $\omega_{x/y/z/w}$ are the components of the quaternion $\boldsymbol{\omega}$. 

\begin{figure}[h]
    \centering
    \input{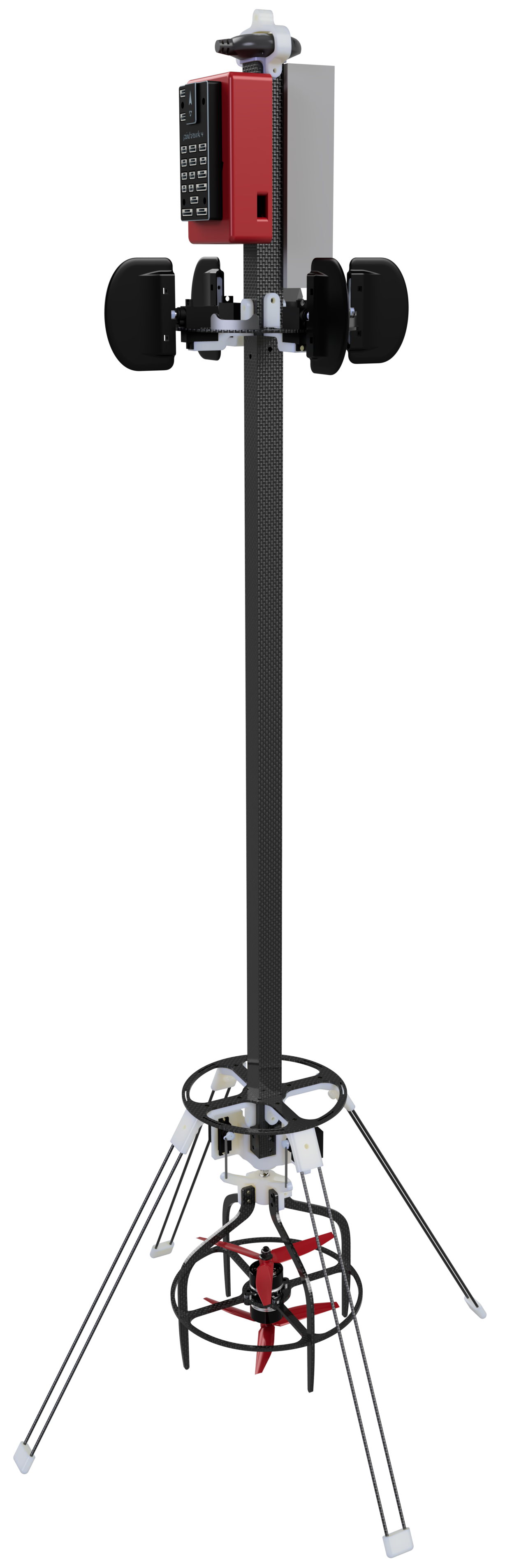}
    \caption{Structural overview of the EmboRockETH: (A) electronics \& battery, (B) 3D-printed fins, (C) carbon fiber central spine, (D) gimbaling mechanism \& propeller cage.}
    \label{fig:structure}
\end{figure}

\section{Vehicle Design}\label{sec:hw}
In this section, we detail the design of the EmboRockETH, which is an electric small-scale VTOL vehicle and has the capability to fly both indoors and outdoors. The EmboRockETH is mainly built from off-the-shelf components with a total cost of less than 1000~USD. The fully assembled vehicle weighs 1.16 kilograms, stands 1.05 meters tall, and can continuously hover for about 5 minutes. In the following, we discuss the overall structural design, the actuators, and the electronics of the vehicle.

\begin{figure*}[t]
\centering
\input{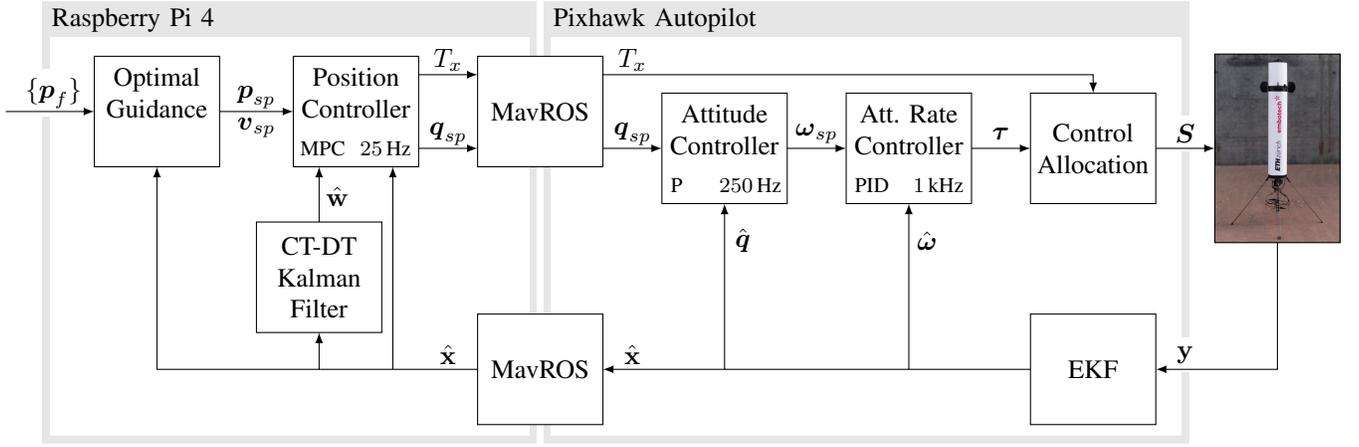}
\caption{The above schematic represents the guidance and control architecture of the EmboRockETH, including the computational execution platform of the individual algorithms. The Raspberry Pi hosts the optimal guidance~\eqref{prob:3DoF_min_fuel} and the high-level MPC position controller~\eqref{eq:mpc}, while the Pixhawk hosts the low-level PID controllers and the control allocation. The communication between the two computation platforms is handled by the ROS package MAVROS.}
\label{fig:architecture} 
\end{figure*}

\subsection{Structural Design}
The EmboRockETH's core structure is a carbon fiber square tube mounted on four flexible non-retractable landing legs made from carbon fiber. The housing of the electronics and the battery are mounted opposite of each other at the top of the central square tube, while the gimbaling mechanism is mounted between the landing legs at the bottom of the vehicle, see Figure~\ref{fig:structure}~(A, C, D). The gimbal consists of an universal joint connecting the central spine to a propeller cage. Additionally, the vehicle is equipped with four 3D-printed fins, which enable aerodynamic control of the vehicle. The four fins are mounted on a ring below the electronics and are symmetrically arranged around the core structure as shown in Figure~\ref{fig:structure}~(B).

\subsection{Actuators}
The EmboRockETH's main propulsion system consists of two coaxially mounted propellers, each actuated by a brushless DC motor. To enable thrust vectoring, the two rotors are mounted in a propeller cage, which is gimbaled by two servo-motors. For aerodynamic steering, the four 3D-printed fins are independently actuated by one servo-motor each.

\subsection{Electronics \& Sensors}
The EmboRockETH uses a Raspberry Pi 4~\cite{RP4} as its main computational platform, while a PixHawk 4 autopilot~\cite{Pixhawk4} provides the sensing capabilities and the low-level flight control. The Raspberry Pi hosts the high-level decision algorithms, i.e., guidance and position control, running on the robot operating system (ROS). The Pixhawk autopilot interfaces the Raspberry Pi through MAVROS~\cite{MAVROS} and hosts the low-level attitude control and state estimator. A more detailed overview of the algorithms and their respective execution platform is shown in Figure~\ref{fig:architecture}. Additionally, the Pixhawk provides all the sensor measurements needed for estimation, i.e. the GPS signal and the inertial measurement unit (IMU) signal. For indoor flights, the GPS module is disabled and replaced by a  motion capture system. This is interfaced through the Raspberry Pi, which relays the measurements to the estimation algorithm on the Pixhawk. The computational platforms and the actuators are powered by a \unit[14.8]{V} lithium polymer battery mounted at the top of the vehicle.

\section{Guidance and Control Architecture}\label{sec:control-architecture}
In this section, we describe the guidance and control architecture of the EmboRockETH that consists of a real-time optimal guidance algorithm, a model predictive control (MPC) position controller, two low-level PID attitude controllers, and a control allocation algorithm. Additionally, the state of the EmboRockETH is estimated using an extended Kalman filter (EKF). The overall architecture is shown in Figure~\ref{fig:architecture} and the individual blocks are discussed in the following sections. 

\subsection{Optimal Guidance}
In the following, we present the optimal guidance used on the EmboRockETH. The guidance leverages previous work presented in~\cite{Jerez2017, Pascucci2019}, but is adapted to the specific properties of the EmboRockETH, i.e. the electric propulsion system and the fixed mass of the vehicle. The optimal guidance problem is formulated as follows
\begin{subequations}\label{prob:3DoF_min_fuel}
    \allowdisplaybreaks
    \begin{align}
    &\min_{t_f, \bvt{T}(\cdot)} \quad \int_{t_0}^{t_f}{\|\bvt{T}(t)\| + \lambda_1 \|\bvt{\dot{T}} (t)\|^2 + \lambda_2 \eta(t)^2  ~ dt} \label{g:min_fuel_cost}\\
	&\text{s.t.}\nonumber \\ 
    &\bv{p}(t_0) = \bv{p}_{0} \qquad \| \bv{p}(t_f) - \bv{p}_{f} \| \leq p_{tol} \label{g:final_pos} \\
	&\bv{v}(t_0) = \bv{v}_{0} \qquad \| \bv{v}(t_f) - \bv{v}_{f} \|  \leq v_{tol} \label{g:final_vel} \\
	&\dot{\bv{p}}(t) = \bv{v}(t) \qquad\! \dot{\bv{v}}(t) = \frac{1}{m} \bvt{T}(t) + \bv{g} \label{g:model} \\
	&\e{1}\!\cdot\!( \bv{p}(t) \!-\! \bv{p}(t_{0})) \geq \tan(\gamma) \,\|\, [\e{2} ~\e{3}] \!\cdot\!(\bv{p}(t) \!-\! \bv{p}(t_{0})) \,\| \label{g:glide_asc} \\
    &\e{1}\!\cdot\!( \bv{p}(t) \!-\! \bv{p}(t_{f})) \geq \tan(\gamma) \,\|\, [\e{2} ~\e{3}] \!\cdot\!(\bv{p}(t) \!-\! \bv{p}(t_{f})) \,\| \label{g:glide_dsc} \\
    &\| \bv{v}(t) \| \leq v_{max} + \eta(t) \qquad \eta(t) \geq 0 \label{g:vel}\\
    &0 \leq T_{min} \leq \| \bvt{T}(t) \| \leq T_{max} \label{g:thrust}\\
	&\dot{T}_{min} \leq \| \bvt{\dot{T}}(t) \| \leq \dot{T}_{max} \label{g:thrust_rate}\\
	&\e{1} \cdot \bvt{T} (t) \geq \cos(\theta_{max}) \| \bvt{T} (t) \| \label{g:tilt}
	\end{align}
\end{subequations}
where the subscripts $0$ and $f$ indicate initial and final values, respectively, $\lambda_1$ and $\lambda_2$ are scalar weighting factors, $\eta(t)$ is a slack variable on the velocity constraint, $p_{tol}$ and $v_{tol}$ are the tolerated deviations from the final position and velocity, respectively, $\gamma$ is the half-angle defining the aperture of the glide slope cone, and $\theta_{max}$ is the maximum tilt angle. The optimal guidance is based on a free-final-time formulation minimizing fuel consumption and violation of the velocity constraint, while discouraging large changes in the input~\eqref{g:min_fuel_cost}. It treats the thrust rate as a control input in order to account for the time response of the actuators. Additionally, it uses a three degrees of freedom~(3-DoF) model~\eqref{g:model} subject to a soft velocity constraint~\eqref{g:vel} and lower and upper bounds on the magnitudes of thrust and thrust rate~\eqref{g:thrust} -~\eqref{g:thrust_rate}. In this 3-DoF formulation, we use the thrust vector as a proxy for the attitude. Therefore, the constraint~\eqref{g:tilt} limits the tilt with respect to the vertical axis in the inertial frame.  The constraints~\eqref{g:glide_asc} -~\eqref{g:glide_dsc} are the ascent and descent glide slope constraints, respectively. During ascent only~\eqref{g:glide_asc} is included in the problem formulation, while~\eqref{g:glide_dsc} is only considered for the descent phase. These two constraints limit the ascent trajectory to a safe corridor and prohibit too shallow descent trajectories.

Problem~\eqref{prob:3DoF_min_fuel} is discretized using the forward Euler method over $30$ points in the interval $[t_0,t_f]$. It is worth noting that, to implement this formulation as free-final-time, the $[t_0,t_f]$ time range is mapped into the $[0, 1]$ pseudo-time interval and the dynamics scaled accordingly. The optimal guidance problem is only solved when a request is received. This occurs shortly before takeoff for the ascent target, if the vehicle is close to the end of the currently computed trajectory, if the tracking error exceeds a predefined threshold, or the desired landing position is changed during the descent phase. 

\subsection{MPC-based Position Controller}
In this section, we describe the MPC-based position controller used for the EmboRockETH. The next sections are organized as follows. First, model and constraints are presented. We then introduce an augmented model that captures the model mismatch and disturbances that can be estimated by a Kalman filter. Finally, we propose the MPC control scheme.
\subsubsection{Model}
To control the position of the EmboRockETH, we leverage a more complex dynamical model~\cite{Szmuk2018} compared to the one used in the optimal guidance. The model is described by the following differential equations
\begin{subequations}
\label{eq:nominal_model}
\begin{align}
\dot{\boldsymbol{p}}(t)      &=  \boldsymbol{v}(t) \\
\dot{\boldsymbol{v}}(t)      &=  \boldsymbol{R}(\boldsymbol{q}(t))\cdot \mathbf{T}(t)/m+\boldsymbol{g} \\
\dot{\boldsymbol{q}}(t)      &=  0.5 \cdot \boldsymbol{Q}(\boldsymbol{\omega}(t))\cdot \boldsymbol{q}(t) \label{eq:nominal_model_quaternion}\\
\dot{\boldsymbol{\omega}}(t) &= \boldsymbol{J}^{-1}\cdot (\mathbf{T}(t) \times \boldsymbol{r}_{G} -\boldsymbol{\omega}(t) \times (\boldsymbol{J} \cdot \boldsymbol{\omega}(t))) \\
\dot{\mathbf{T}}(t)      &= (\boldsymbol{u}(t)-\mathbf{T}(t))/t_{\tau}, \label{eq:nominal_model_thrust_dynamics}
\end{align}
\end{subequations}
where the state of the system is given by ${\mathbf{x} =  \left[\boldsymbol{p}^\top, \boldsymbol{v}^\top, \boldsymbol{q}^\top, \boldsymbol{\omega}^\top, \mathbf{T}^\top \right]^\top}$. Equation~\eqref{eq:nominal_model_quaternion} is norm preserving, implying that $\|q\|=1$  The model's input is the desired thrust vector and it is denoted by $\boldsymbol{u} \in \mathbb{R}^3$. To model the delay in response of the motors, the actual thrust dynamics $\mathbf{T}$ is modeled as a first order system with time constant $t_\tau \in \mathbb{R}$, as described in equation~\eqref{eq:nominal_model_thrust_dynamics}. 

The distance between the engine's hinge point and the vehicle's center of gravity is represented by $\boldsymbol{r}_{G} = [r_G,0,0]^\top \in \mathbb{R}^{3}$, while the inertia matrix is denoted by $\boldsymbol{J} \in \mathbb{R}^{3\times 3}$. The nominal system dynamics can be compactly rewritten as $\dot{\boldsymbol{x}}(t) = f(\boldsymbol{x}(t),\boldsymbol{u}(t))$.

\subsubsection{Disturbance Model and State Estimation}
In order to consider partial state knowledge, sensor noise, model mismatch and unknown disturbances, the nominal model is augmented with an output model and an uncertainty term in the state-space equation.  The uncertain system dynamics can then be written as
\begin{equation}
\label{eq:model_with_disturbance}
\begin{split}
    &\dot{\mathbf{x}}(t) = f(\mathbf{x}(t), \mathbf{u}(t)) + \mathbf{B_d} \mathbf{w}(\mathbf{x}(t), \mathbf{u}(t)),\\
    &\dot{\mathbf{w}}(t)(\mathbf{x}(t), \mathbf{u}(t)) = 0,\\
    &\mathbf{y}(t) = \mathbf{C}\mathbf{x}(t) + \mathbf{v}(t),
\end{split}
\end{equation}
where we assume that the uncertainty in~\eqref{eq:model_with_disturbance} has zero-order dynamics~\cite{FBMM:07}, i.e., the disturbance is constant, $\mathbf{B_d}$ is the noise to state matrix, $\mathbf{w}(\mathbf{x}, \mathbf{u})$ models the unknown uncertainty and disturbances, $\mathbf{C}$ is the output matrix, and $\mathbf{v}$ is the measurement noise, which is assumed to be independent of the state or input and Gaussian i.i.d. with zero mean and covariance~$\Sigma_v = \sigma^2\mathbb{I}$. We further model the uncertainty term such that $\mathbf{w}(\mathbf{x},\mathbf{u}) = \left[ d\boldsymbol{v}^\top, d\boldsymbol{a}^\top, d\boldsymbol{\alpha}^\top \right]^\top$, where $d\boldsymbol{v} = \left[ dv_x, dv_y, dv_z \right]^\top$ is the mismatch between the actual and nominal velocity, $d\boldsymbol{a} = \left[ da_x, da_y, da_z \right]^\top$ is the mismatch between the actual and nominal acceleration, and 
$d\boldsymbol{\alpha} = \left[ d\alpha_x, d\alpha_y, d\alpha_z \right]^\top$ is the mismatch between the actual and nominal angular acceleration. The system dynamics in~\eqref{eq:model_with_disturbance} are discretized using the Runge-Kutta 4 method with $4$ intermediate nodes at with a sampling-time of $25Hz$, obtaining a non-linear system of the form
\begin{equation}
\label{eq:RK4model}
\begin{split}
&\mathbf{x}(k+1) = \bar{f}(\mathbf{x}(k), \mathbf{u}(k),\mathbf{w}(k)),\\
&\mathbf{w}(k+1) = \mathbf{w}(k),\\
&\mathbf{y}(k) = \mathbf{C} \mathbf{x}(k) + \mathbf{v}(k),
\end{split}
\end{equation}
where $\bar{f}$ are the discretized system dynamics.
As we directly measure position, asset, acceleration, and angular acceleration, we assume model~\eqref{eq:RK4model} to be observable, such that the state~$\mathbf{x}$ and offset $\mathbf{w}$ can be estimated using an extended Kalman filter and a CT-DT Kalman filter~\cite{RKRB:61}, respectively.

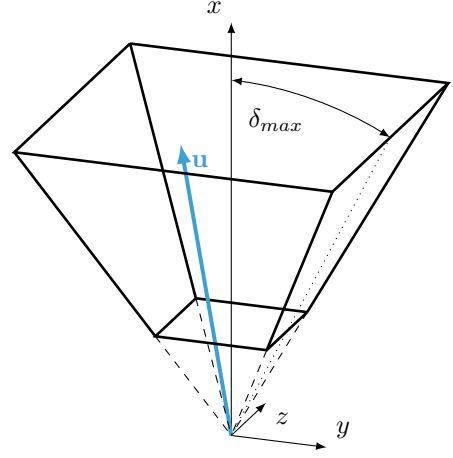
\begin{figure}[t]
\centering
\tdplotsetmaincoords{70}{110}
\begin{tikzpicture}[>=latex,tdplot_main_coords, scale=0.9]
\pgfmathsetmacro{\clen}{1.5}
\pgfmathsetmacro{\xlen}{6.5}
\pgfmathsetmacro{\hpyr}{-5}
\pgfmathsetmacro{\hpyrtop}{-1.75}
\pgfmathsetmacro{\apyr}{2.5}
\pgfmathsetmacro{\apyrtop}{\apyr*\hpyrtop/\hpyr}

\definecolor{mycolor}{rgb}{0.23529,0.62745,0.82353}

\draw[->] (0,0,0) -- (-\clen,0,0) node[anchor=north west]{$z$};
\draw[->] (0,0,0) -- (0,\clen,0) node[anchor=south west]{$y$};
\draw[->] (0,0,0) -- (0,0,\xlen) node[anchor=south east]{$x$};

\draw[line width = 1] (-\apyrtop,-\apyrtop,-\hpyrtop) -- (-\apyrtop,\apyrtop,-\hpyrtop);
\draw[->,mycolor,line width = 1.5] (0,0,0) -- (-0.5*\apyr,-0.5*\apyr,-0.8*\hpyr)node[anchor=north west]{$\mathbf{u}$};

\draw[line width = 1] (-\apyr,-\apyr,-\hpyr) -- (\apyr,-\apyr,-\hpyr) -- (\apyr,\apyr,-\hpyr) -- (-\apyr,\apyr,-\hpyr) -- (-\apyr,-\apyr,-\hpyr);

\draw[line width = 1] (-\apyrtop,-\apyrtop,-\hpyrtop) -- (\apyrtop,-\apyrtop,-\hpyrtop) -- (\apyrtop,\apyrtop,-\hpyrtop) -- (-\apyrtop,\apyrtop,-\hpyrtop);

\draw[line width = 1] (-\apyr,-\apyr,-\hpyr) -- (-\apyrtop,-\apyrtop,-\hpyrtop);
\draw[line width = 1] (\apyr,-\apyr,-\hpyr) -- (\apyrtop,-\apyrtop,-\hpyrtop);
\draw[line width = 1] (\apyr,\apyr,-\hpyr) -- (\apyrtop,\apyrtop,-\hpyrtop);
\draw[line width = 1] (-\apyr,\apyr,-\hpyr) -- (-\apyrtop,\apyrtop,-\hpyrtop);

\draw[dashed] (-\apyrtop,-\apyrtop,-\hpyrtop) -- (0,0,0);
\draw[dashed] (\apyrtop,-\apyrtop,-\hpyrtop) -- (0,0,0);
\draw[dashed] (\apyrtop,\apyrtop,-\hpyrtop) -- (0,0,0);
\draw[dashed] (-\apyrtop,\apyrtop,-\hpyrtop) -- (0,0,0);

\draw[dotted] (0,\apyr,-\hpyr) -- (0,0,0);
\tdplotsetthetaplanecoords{90}
\tdplotdrawarc[tdplot_rotated_coords,<->]{(0,0,0)}{{sqrt(\apyr*\apyr+\hpyr*\hpyr)}}{0}{-atan(\apyr/\hpyr)}{anchor=north east}{$\delta_{max}$}

\end{tikzpicture}
\caption{Polytopic set defined by the input constraints~\eqref{eq:input_constraints} including an example input vector $\boldsymbol{u}$ (in blue) satisfying the constraints.}
\label{fig:const} 
\end{figure}

\subsubsection{Constraints}
The system is subject to the following polytopic input constraints 
\begin{subequations}\label{eq:input_constraints}
\begin{alignat}{2}
-u_{x}\tan(\delta_{max})&\leq u_{y}&&\leq u_{x}\tan(\delta_{max}), \\
-u_{x}\tan(\delta_{max})&\leq u_{z}&&\leq u_{x}\tan(\delta_{max}), \\
u_{x,min}          &\leq u_{x}             &&\leq u_{x,max},
\end{alignat}
\end{subequations}
where $\delta_{max}$ is the maximum gimbal angle. The set formed by~\eqref{eq:input_constraints} is depicted in Figure~\ref{fig:const}. As the thrust constraints of the propulsion system are defined by the minimal and maximal achievable thrust magnitudes $\| T \|_{min}$ and $\| T \|_{max}$, respectively, we need to convert these constraints into $u_{x,min}$ and $u_{x,max}$, i.e.,
\begin{subequations} \label{eq:constriant_conversion}
\begin{align}
    u_{x,min} &= || \mathbf{T} ||_{min}, \\
    u_{x,max} &= \frac{|| \mathbf{T}  ||_{max}}{\sqrt{1+2\tan^2(\delta_{max})}}.
\end{align}
\end{subequations}
The conversion equations~\eqref{eq:constriant_conversion} define the lower and upper bounds of a pyramid, which is inscribed in the spherical sector defined by $\| T \|_{min}$ and $\| T \|_{max}$. We only consider input constraints, therefore our MPC formulation does not include state constraints.

\subsubsection{Offset-free Tracking MPC}
We propose a non-linear MPC control scheme for system~\eqref{eq:RK4model}, which exploits the disturbance model to perform offset-free tracking and ensures constraint satisfaction. In the following, we will assume that the optimal guidance and the offset-free tracking MPC controller are time-synchronized. The following optimization problem defines the MPC scheme
\begin{subequations}\label{eq:mpc}
\begin{align}
    \min_{\mathbf{x}_{i|k},\mathbf{u}_{i|k}} & l_f(\mathbf{x}_{N|k},\mathbf{r}_{N|k})  +\sum_{i = 0}^{N-1}l_i(\mathbf{x}_{i|k},\mathbf{u}_{i|k},\mathbf{w}_{i|k},\mathbf{r}_{i|k}) \\
    \text{s.t.~} & \forall i \in \{0,1,\ldots,N-1\}\\
    & \mathbf{x}_{0|k} = \mathbf{\hat{x}}(k)\\
    & \mathbf{w}_{0|k} = \mathbf{\hat{w}}(k)\\
    & \mathbf{x}_{i+1|k} = \bar{f}(\mathbf{x}_{i|k},\mathbf{u}_{i|k},\mathbf{w}_{i|k}) \label{eq:mpc_dynamics}\\
    &\mathbf{w}_{i+1|k} = \mathbf{w}_{i|k}\label{eq:mpc_offset}\\
    & \mathbf{A}\mathbf{u}_{i|k} \leq \mathbf{b}, \label{eq:mpc_constraints}
\end{align}
\end{subequations}
where the subscript $_{i|k}$ denotes the $i$-th prediction starting from the initial condition at the time instant $k$. The vector $\mathbf{r}=[\mathbf{p}_{sp}^\top,  \boldsymbol{v}_{sp}^\top]^\top$ is the position and velocity reference provided by the optimal guidance and is kept constant over the whole horizon, $\mathbf{\hat{x}}(k)$ and $\mathbf{\hat{w}}(k)$ are the initial conditions of the optimization problem obtained by the Kalman filters, and $N$ is the prediction horizon. 
The terms $l_i$ and $l_f$ are appropriate functions for reference tracking, for example quadratic functions that penalize the error between state and reference, i.e.,
\begin{align}
\label{eq:stage_cost}
\begin{split}
l_i = &||\boldsymbol{p}_i-\boldsymbol{p}_{sp}||^2_{Q_p} + ||(\boldsymbol{v}_i+d\boldsymbol{v})-\boldsymbol{v}_{sp}||^2_{Q_v} + \\ &||\boldsymbol{\omega}_{yz,i}||^2_{Q_\omega} + ||\mathbf{T}_i-\mathbf{T}_{eq}||^2_{Q_T} + ||\boldsymbol{u}_i-\mathbf{T}_{eq}||^2_R
\end{split}
\end{align}
and $l_f = ||\boldsymbol{p}_N-\boldsymbol{p}_{sp}||^2_{Q_N}$ with
\begin{align}
\label{eq:thrust_equilibrium}
\boldsymbol{\omega}_{yz,i} = \begin{bmatrix} \omega_{y,i} \\ \omega_{z,i} \end{bmatrix},\ 
\mathbf{T}_{eq} = \begin{bmatrix} m\cdot (g-d\boldsymbol{a}_x) \\ (J_{zz}\cdot d\boldsymbol{\alpha}_z)/r_G \\(J_{yy}\cdot d\boldsymbol{\alpha}_y)/r_G \end{bmatrix}.
\end{align}
The first term in~\eqref{eq:stage_cost} accounts for the deviation between the predicted and desired trajectory. The second penalizes the deviation between the desired velocity and the offset-free predicted velocity. The third term helps in dampening the oscillations produced by large variations of the angular accelerations in the $y$ and $z$-axes. The last two terms penalize the deviation of the thrust from the desired offset-free thrust equilibrium~$\mathbf{T}_{eq}$, computed as in~\eqref{eq:thrust_equilibrium}.
Finally, equations~\eqref{eq:mpc_dynamics}-\eqref{eq:mpc_offset} represent the system dynamics and equation~\eqref{eq:mpc_constraints} the input constraints~\eqref{eq:input_constraints}. 
Problem~\ref{eq:mpc} can be efficiently solved using nonlinear programming solvers such as~\cite{FORCESNLP}. 

\label{sec:CA}
\begin{figure}[t]
\centering
\input{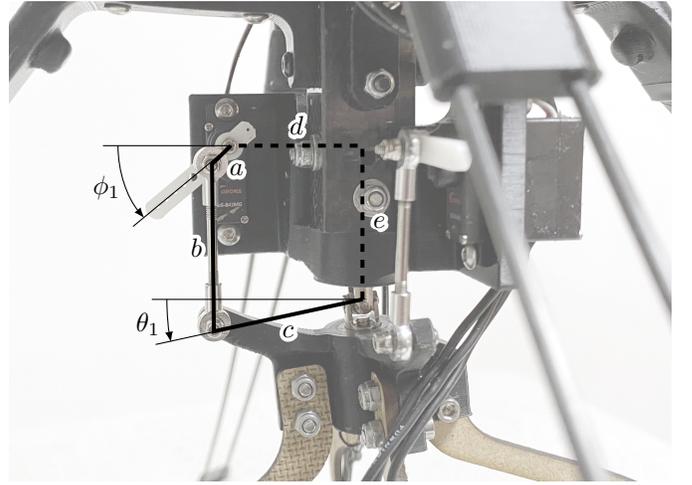}
\caption{Above picture shows the gimbal and a schematic of its geometry for one gimbaling direction. The gimbal's deflection in this direction is determined by the position of servo arm (a) and the geometry. The servo-motor rotates the servo arm (a) by the angle $\phi_1$. Given the geometry of the gimbal, this motion rotates the gimbal arm (c) by the angle $\theta_1$, which then defines the deflection of the gimbal. The symbols $a,b,c,d,$ and $e$ denote the names and lengths of the five arms defining the gimbal's geometry.}
\label{fig:3Dkin} 
\end{figure}

\subsection{Low-Level PID Controllers}
The attitude and attitude rate PID controllers run on the Pixhawk autopilot. As they are computationally lightweight, they are executed at high sampling rates, i.e., $250$Hz and $1$kHz, respectively. The input to the low-level attitude controller is the quaternion set-point $\mathbf{q}_{sp}$ generated by the MPC controller and its output is the velocity set-point~$\boldsymbol{\omega}_{sp}$ for the attitude rate controller. This in turn outputs the torque~$\boldsymbol{\tau}$ that should be generated by the actuators. This structure reflects the architecture presented in~\cite{Brescianini2013}.

\begin{figure*}[t]
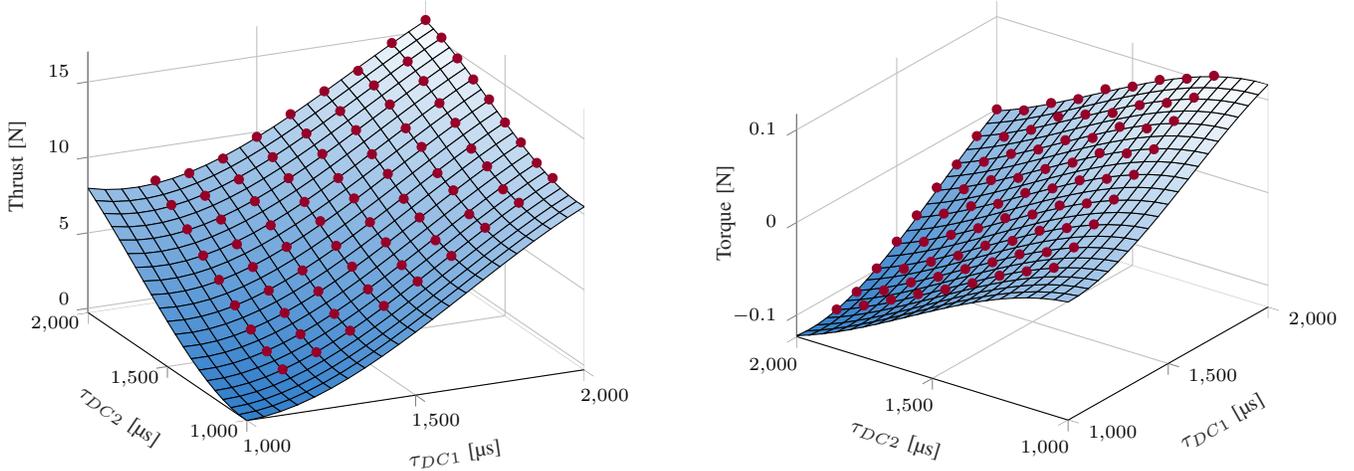

\begin{minipage}[c]{0.5\linewidth}
\centering
\vspace{0.1cm}
\input{media/thrust_ID.tex}
\end{minipage}
\begin{minipage}[c]{0.55\linewidth}
\centering
\input{media/torque_ID.tex}
\end{minipage}
\caption{The experimentally determined thrust magnitude $||\mathbf{T}||$ (left) and torque $\tau_x$ (right) as a function of the moter PWM's duty cycle commands. The red dots indicate thrust and torque measurements averaged over a \unit[2]{s} horizon and the blue surfaces show the cubic surface fits, which are used as models.}
\label{fig:mapping_id} 
\end{figure*}

\subsection{Control Allocation}
The control allocation converts the torque $\boldsymbol{\tau} = [\tau_x, \tau_y, \tau_z]^\top$ and thrust along the body x-axis $T_x$ into commands for the actuators. The controlled actuators are the two servo-motors, which determine the orientation of the universal joint, and the two DC motors connected to the propellers. The vehicle's fins are currently not considered in the control allocation and will be the subject of future research. The allocation algorithm is described in Algorithm~\ref{alg:control_allocation}. 
At first, the control allocation maps the torque $\boldsymbol{\tau}$ and the thrust $T_x$ into an equivalent three-dimensional thrust vector acting on the center of the universal joint, i.e.
\begin{equation}
\label{eq:thrust_vector}
\mathbf{T} = \begin{bmatrix}T_x\\T_y\\T_z\end{bmatrix} = \begin{bmatrix}T_x\\-\tau_z/r_G\\ \tau_y/r_G\end{bmatrix}.
\end{equation}
The second step consists of determining the gimbal angles given the thrust vector. 
As depicted in Figure~\ref{fig:3Dkin}, the gimbal angles $\theta_1$ and $\theta_2$ correspond to the thrust vector orientation with respect to the center of the universal joint, and their relationship is given by the following equalities
\begin{equation}
\label{eq:thrust_vector_angels}
\begin{split}
    \theta_1 &= -\arcsin \left( \frac{T_z}{||\mathbf{T}||_2-T_y^2}\right)\\
    \theta_2 &= \hphantom{-}\arcsin \left( \frac{T_y}{||\mathbf{T}||_2}\right).
    \end{split}
\end{equation}
Next, the kinematics of the gimbal is inverted to obtain the servo angles $\phi_1$ and $\phi_2$, see Figure \ref{fig:3Dkin}. The first servo-motor rotates the gimbal around the y-axis with angle $\phi_1$ and defines the axis of rotation for the second servo-motor. From the geometry of the gimbaling mechanism, it is possible to derive the two following equalities
\begin{equation}
\label{eq:servo_motor_angles}
\begin{split}
b^2 = &(e - a\sin(\phi_1) + c\sin(\theta_1))^2 + \\
                        &(d - c\cos(\theta_1) + a\cos(\phi_1))^2, \\
b^2 = &(e - a\sin(\phi_2) + c\cos(\theta_1)\sin(\theta_2))^2 + \\
                        &(d - c\cos(\theta_2) + a\cos(\phi_2))^2 + \\
                        &c^2\sin(\theta_2)^2\sin(\theta_1)^2.
\end{split}
\end{equation}
These equations can be solved explicitly to obtain~$\phi_1$ and~$\phi_2$. Finally, the last step consists of finding the motor commands~$\tau_{DC1}$ and~$\tau_{DC2}$ such that the torque $\tau_x$ and the thrust $||\mathbf{T}||$ are generated by the propeller pair. These two maps can be described by two third-order polynomials, see Figure~\ref{fig:mapping_id}. In Section~\ref{sec:loadcellexp}, we detail how these maps have been experimentally determined.

\begin{algorithm}
\caption{Control Allocation}\label{alg:control_allocation}
\hspace*{\algorithmicindent} \textbf{Input} Torque $\boldsymbol{\tau}$, Thrust  $T_x$  \\
\hspace*{\algorithmicindent} \textbf{Output} $\phi_1$, $\phi_2$, $\tau_{DC1}$, $\tau_{DC2}$

\begin{algorithmic}[1]
\STATE Computation of the thrust vector using~\eqref{eq:thrust_vector}
\STATE Computation of the gimbal angles using~\eqref{eq:thrust_vector_angels}
\STATE Solve~\eqref{eq:servo_motor_angles} to obtain the servo-motor angles~$\phi_1$ and~$\phi_2$
\STATE Compute PWM duty cycle motor commands~$\tau_{DC1}$ and~$\tau_{DC2}$ from~$\tau_x$ and~$||\mathbf{T}||$
\end{algorithmic}
\end{algorithm}

\section{Experimental Results}\label{sec:experiments}
We performed three kinds of experiments. The first experiment identified the motor command maps. These maps are necessary to covert thrust and torque commands into motor commands. The second experiment tested the offset-free tracking MPC using a reliable and accurate localization system, i.e. a motion capture system. Finally, we tested the overall guidance and control pipeline using the on-board GPS as localization system. All the experiments were conducted using the FORCESPRO~\cite{FORCESNLP, FORCESPro} non-linear programming solver to solve the optimization problems~\eqref{prob:3DoF_min_fuel} and~\eqref{eq:mpc} on the Raspberry Pi (\unit[1.8]{GHz} ARM Cortex A72, \unit[4]{GB} memory) in real-time. In the Raspberry Pi, two cores were used to run the guidance and one to run the MPC controller. 

\subsection{Identification of Motor Command Maps}
\label{sec:loadcellexp}
In the final step of Algorithm~\ref{alg:control_allocation}, the thrust magnitude~$||\mathbf{T}||$ and the torque~$\tau_x$ have to be mapped to the motor commands~$\tau_{DC1}$ and~$\tau_{DC2}$, which represent the pulse width modulation's duty cycle, where the fundamental frequency is set to~\unit[400]{Hz}. These maps have to be determined experimentally, since they vary depending on the specific motors and propeller configuration used. In the following, we detail the experimental procedure that obtains these maps. While individually increasing the motor commands~$\tau_{DC1}$ and~$\tau_{DC2}$ in \unit[100]{$\mu$s} increments, we measure the thrust~$||\mathbf{T}||$ and torque~$\tau_x$ with a load cell. Subsequently, we fit two cubic surfaces to the thrust and torque measurements, see Figure~\ref{fig:mapping_id}. Then, the two maps are inverted to obtain a description of the motor commands as a function of the thrust and torque. These inversions are always feasible since the cubic surfaces are required to be strictly monotonic in the admissible range of motor commands. Since the measurements are performed at the nominal voltage of the battery, deviations arise due to variations in the battery voltage. These deviations are compensated by the offset-free MPC formulation and are not accounted for in the control allocation.

\begin{figure}[ht]
    \centering
    \input{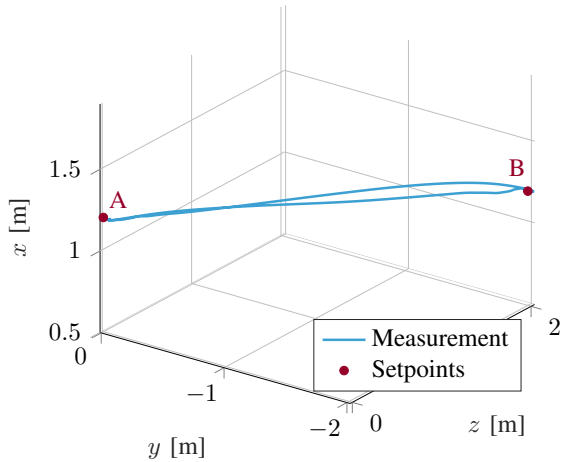}
    \caption{The measured indoor flight trajectory (blue line) between set-points A and B is shown for a control configuration as in Fig.~\ref{fig:architecture}, but excluding the optimal guidance. The vehicle starts in A $\mathbf{x} = [ \unit[1.2]{m}, \unit[0]{m}, \unit[0]{m}]^\top$, is then commanded to fly to B $\mathbf{x} = [ \unit[1.2]{m}, \unit[2]{m}, \unit[2]{m}]^\top$, before returning to A.}
    \label{fig:3d_MPC}
\end{figure}

\subsection{Indoor Flight for Validation}
\begin{figure}[t]
    \centering
    \input{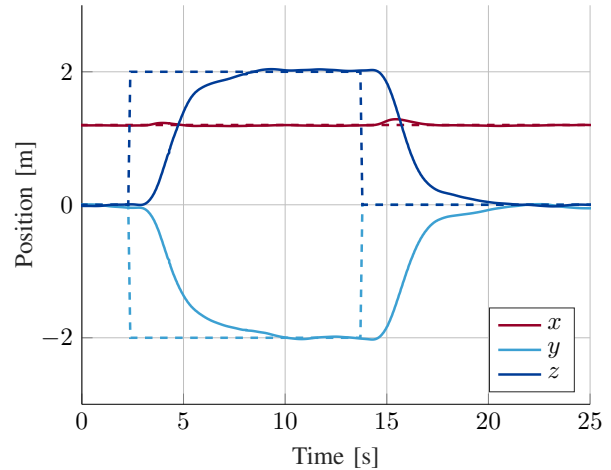}
    \caption{Behavior of the offset-free MPC position controller in the indoor environment to a step input in the $y-$ and $z-$direction. The position along the $x$-, $y$- and $z$-axis is shown by the solid lines and the dashed lines indicate the corresponding set-points.}
    \label{fig:MPC_pos}
\end{figure}

\begin{figure}[ht]
    \centering
    \input{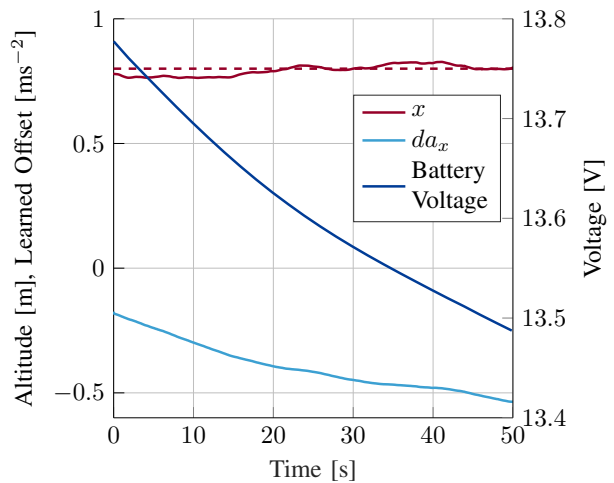}
    \caption{The plot shows the vehicle tracking a constant reference (dashed red line). The measured vehicle trajectory (solid red line) is close to the reference for the whole experiment even if there is a battery voltage drop (dark blue line) of \unit[0.3]{V}. This is achieved by the CT-DT Kalman filter through continuously adapting the estimated model offset (light blue line).}
    \label{fig:bat_volt}
\end{figure}

In a first experiment, we test the performance of the position control-loop, i.e., the architecture in Figure~\ref{fig:architecture} without the optimal guidance. The experiment is performed indoors and the vehicle is localized using a Vicon motion capture system. The reference and the vehicle trajectory are shown in Figure~\ref{fig:3d_MPC} for a flight between the set-points A and B. Additionally, Figure~\ref{fig:MPC_pos} shows the references and trajectories for all three axes individually. The MPC offset-free tracking formulation compensates the model mismatch, given e.g. by the misalignment between the center of gravity of the vehicle and the barycenter of the central rod, resulting in the offset-free tracking observed in Figures~\ref{fig:3d_MPC} and~\ref{fig:MPC_pos}. In a second experiment, we further highlight the effectiveness of the offset-free formulation. We let the EmboRockETH hover at a fixed position and observe if it can hold the altitude as the battery voltage drops. The results are shown in Figure~\ref{fig:bat_volt}, notice the voltage drop and the corresponding adjustment of the estimated uncertainty term $da_x$ in order to keep the vehicle at the desired position.

\subsection{Outdoor Flight}
\begin{figure}[ht]
    \centering
    \input{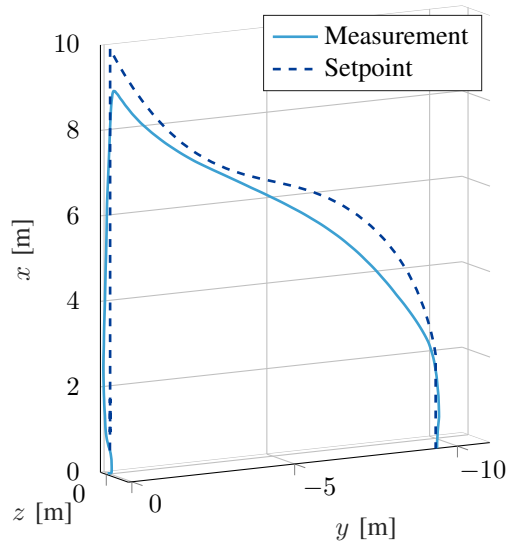}
    \caption{Trajectory of a mission (dashed line) generated in real-time by the optimal guidance~\eqref{prob:3DoF_min_fuel} together with the measured trajectory followed by the vehicle controller (solid line) during an outdoor flight. \label{fig:MPC_GNC_out}}
\end{figure}

\begin{figure}[ht]
    \centering
    \scalebox{-1}[1]{\includegraphics[width=\linewidth]{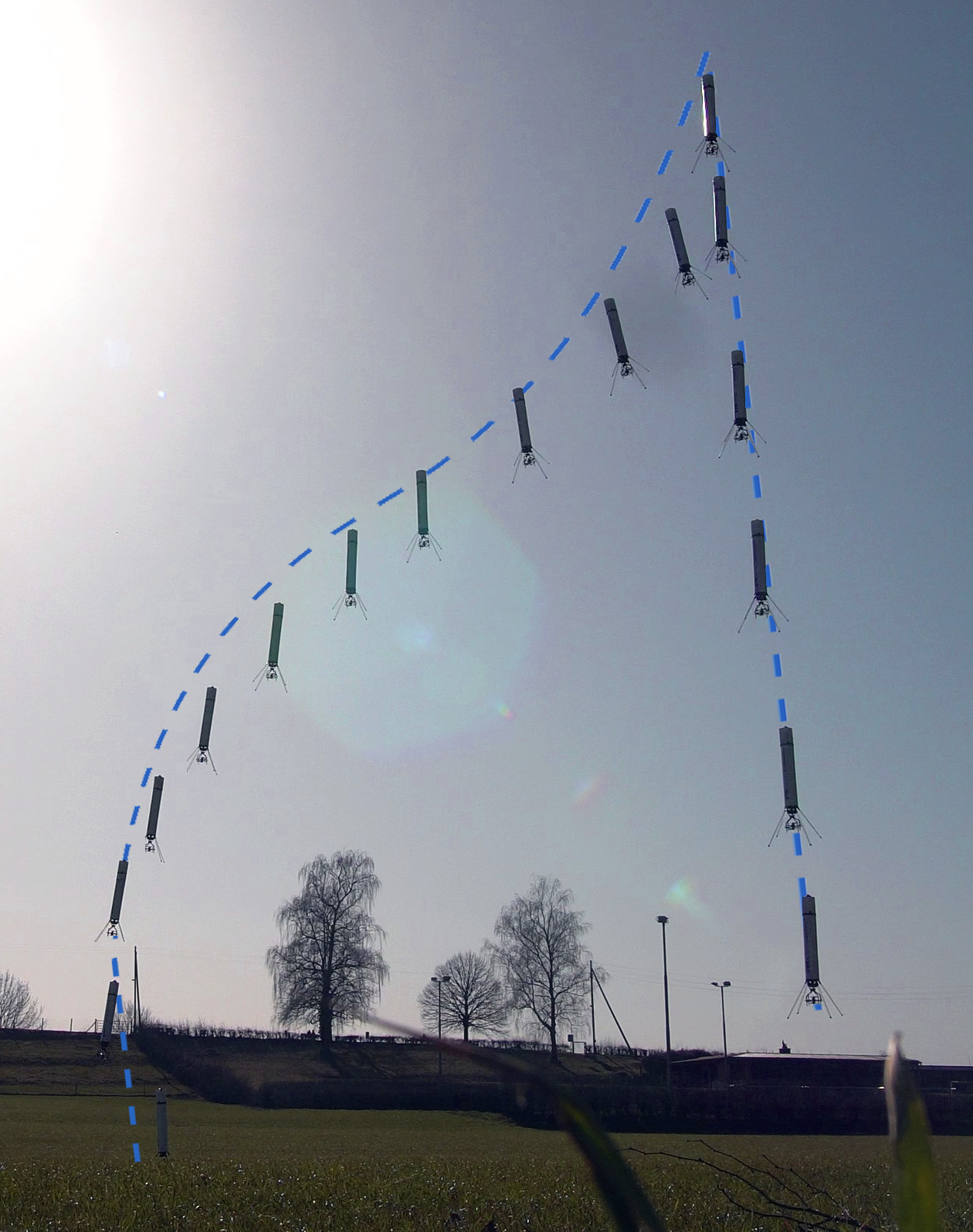}}
    \caption{EmboRockETH in several states along the measured trajectory shown in Fig.~\ref{fig:MPC_GNC_out} and the corresponding set-points.}
    \label{fig:MPC_GNC_vid}
\end{figure}

In this experiment, we demonstrate the performance of the entire control pipeline shown in Figure~\ref{fig:architecture}. Both optimal guidance~\eqref{prob:3DoF_min_fuel} and offset-free MPC~\eqref{eq:mpc} run in real-time on the Raspberry PI with an average solve time of \unit[120]{ms} and \unit[30]{ms}, respectively.  The experiment is performed outdoors and the vehicle is localized using the GPS in combination with the IMU. The mission scenario requires the vehicle to fly to an altitude of ten meters, followed by a combined translation and descent maneuver to the landing pad, see Figure~\ref{fig:MPC_GNC_out}. The maximum velocity reached during the maneuver is \unit[3]{m/s}. Figures~\ref{fig:MPC_GNC_out} and \ref{fig:MPC_GNC_vid} show that the vehicle is able to complete the mission and land within~\unit[50]{cm} of the target.

\section{Conclusions}\label{sec:conclusion}
This paper has proposed an inexpensive small-scale vertical take-off and landing (VTOL) aircraft dubbed \emph{EmboRockETH}, with the purpose to serve as a test-bed for advanced real-time guidance and control algorithms. The EmboRockETH was designed to better capture the dynamical aspects of a re-usable rocket than existing quadrotor platforms and perform indoor and outdoor flights. Additionally, a real-time guidance and control pipeline was proposed for the EmboRockETH and verified in indoor and outdoor experiments.

\addtolength{\textheight}{-1.5cm}   

\bibliographystyle{IEEEtran}
\bibliography{IEEEabrv, bibliography}

\end{document}